\documentclass[lettersize,journal]{IEEEtran}
\usepackage{cite}
\usepackage{amsmath,amssymb,amsfonts}
\usepackage{algorithmic}
\usepackage{graphicx}
\usepackage{array} 
\usepackage{adjustbox}
\usepackage{longtable}
\usepackage{textcomp}
\usepackage{xcolor}
\usepackage{mathrsfs}
\usepackage{caption}
\definecolor{darkgreen}{RGB}{0, 121, 0}
\definecolor{darkred}{RGB}{139, 0, 0}
\usepackage{subcaption} 
\usepackage{url}
\usepackage{enumitem} 
\usepackage[normalem]{ulem}
\usepackage{soul}
\setlist[itemize]{left=0pt} 
\begin{document}

\title{\textbf{MuPlon}: \textbf{Mu}lti-\textbf{P}ath Causa\textbf{l} Optimizati\textbf{on} for \\
Claim Verification through Controlling Confounding
}

\author{\IEEEauthorblockN{Hanghui Guo\IEEEauthorrefmark{1},
Shimin Di\IEEEauthorrefmark{2},
Pasquale De Meo\IEEEauthorrefmark{3},
Zhangze Chen\IEEEauthorrefmark{1}, and
Jia Zhu\IEEEauthorrefmark{1}}\\
\IEEEauthorblockA{\IEEEauthorrefmark{1}Zhejiang Key Laboratory of Intelligent Education Technology and Application,\\
Zhejiang Normal University}\\
\IEEEauthorblockA{\IEEEauthorrefmark{2}School of Computer Science and Engineering, Southeast University}\\
\IEEEauthorblockA{\IEEEauthorrefmark{3}Department of Computer Science, University of Messina}\\
}


\markboth{Journal of \LaTeX\ Class Files,~Vol.~14, No.~8, August~2021}%
{Shell \MakeLowercase{\textit{et al.}}: A Sample Article Using IEEEtran.cls for IEEE Journals}


\maketitle

\begin{abstract}
As a critical task in data quality control,
claim verification aims to 
curb the spread of misinformation by 
assessing the truthfulness of claims based on a wide range of evidence.
However, traditional methods often overlook the complex interactions between evidence, leading to unreliable verification results. A straightforward solution represents the claim and evidence as a fully connected graph, which we define as the Claim-Evidence Graph (C-E Graph). 
Nevertheless, claim verification methods based on fully connected graphs face two primary confounding challenges, \textit{Data Noise} and \textit{Data Biases}. 
To address these challenges, we propose a novel framework, \textit{\textbf{Mu}lti-\textbf{P}ath Causa\textbf{l} Optimizati\textbf{on} \textbf{(MuPlon)}}. 
MuPlon integrates a dual causal intervention strategy, consisting of the \textit{back-door path} and \textit{front-door path}. In the \textit{back-door path}, MuPlon dilutes noisy node interference by optimizing node probability weights, while simultaneously strengthening the connections between relevant evidence nodes. In the \textit{front-door path}, MuPlon extracts highly relevant subgraphs and constructs reasoning paths, further applying counterfactual reasoning to eliminate data biases within these paths. The experimental results demonstrate that MuPlon outperforms existing methods and achieves state-of-the-art performance. 
\end{abstract}


\section{Introduction}

Today, the internet is overwhelmed with a vast array of unverified claims, including fake news, malicious rumors, and misleading advertisements, all of which have significant impacts on politics, economics, law, and other sectors, and may even pose serious threats to public safety \cite{allcott2017social,guo2025dior,zhu2025radio}. Therefore, addressing these challenges requires effective methods to evaluate and ensure the accuracy of information.

One effective method to address these challenges is claim verification, which aims to improve the quality of Internet content by filtering out falsehoods and emphasizing reliable information. This process can effectively reduce the spread of misinformation \cite{guo2025consistent}. Consequently, an increasing number of researchers have concentrated on modeling the claim verification task, with a specific focus on assessing the truthfulness of a given claim by analyzing both the claim itself and its associated evidence \cite{wadden2020fact, west2021misinformation}.


Most existing methods rely on single evidence for independent verification. 
For instance, the ESIM model \cite{chen2016enhanced} leverages sequence reasoning and attention mechanisms to capture the relationship between a claim and a single given evidence, and verify it by calculating a score. Similarly, the KIM model \cite{chen2017neural} identifies core terms from a single evidence and retrieves corresponding synonyms, antonyms, and hypernyms from a database, then combines the core terms from the information and original evidence together through soft alignment and semantic fusion to understand the logical relationship. However, this reliance on single evidence has significant limitations, as it is susceptible to inducements influence from a single source, which can result in a lack of comprehensiveness and diversity in the verification process, ultimately leading to conclusions that lack sufficient credibility and reliability \cite{everitt2021agent}. 

In contrast, integrating multiple pieces of evidence provides a more holistic perspective, which can improve the trustworthiness of the verification. For example,
the Athene model \cite{hanselowski2018ukp} upgrades the ESIM model, which retrieves relevant articles to extract evidence from Wikipedia, calculating scores of claim-evidence pairs using ESIM for ranking, and selecting the top five as the potential evidence set for verification. 
Based on the Athene model, the NSMN model~\cite{nie2019combining} further defines 10 independent channels through vocabulary items in WordNet. 
Each channel is responsible for processing different semantic features. When a word in the evidence is related to a vocabulary item in WordNet (such as a synonym set), the relevant channel is activated and the feature information of the channel is integrated with evidence into the input of the model.
However, traditional multi-evidence verification frameworks simply increase the evidence volume or additional semantic features without effectively capturing the complex interactions between the pieces of evidence.
Neglecting the relationships among evidence may reduce the accuracy of the verification process
and validity of the inference results.

\begin{figure}[!t]
    \centering
    \includegraphics[width=\columnwidth]{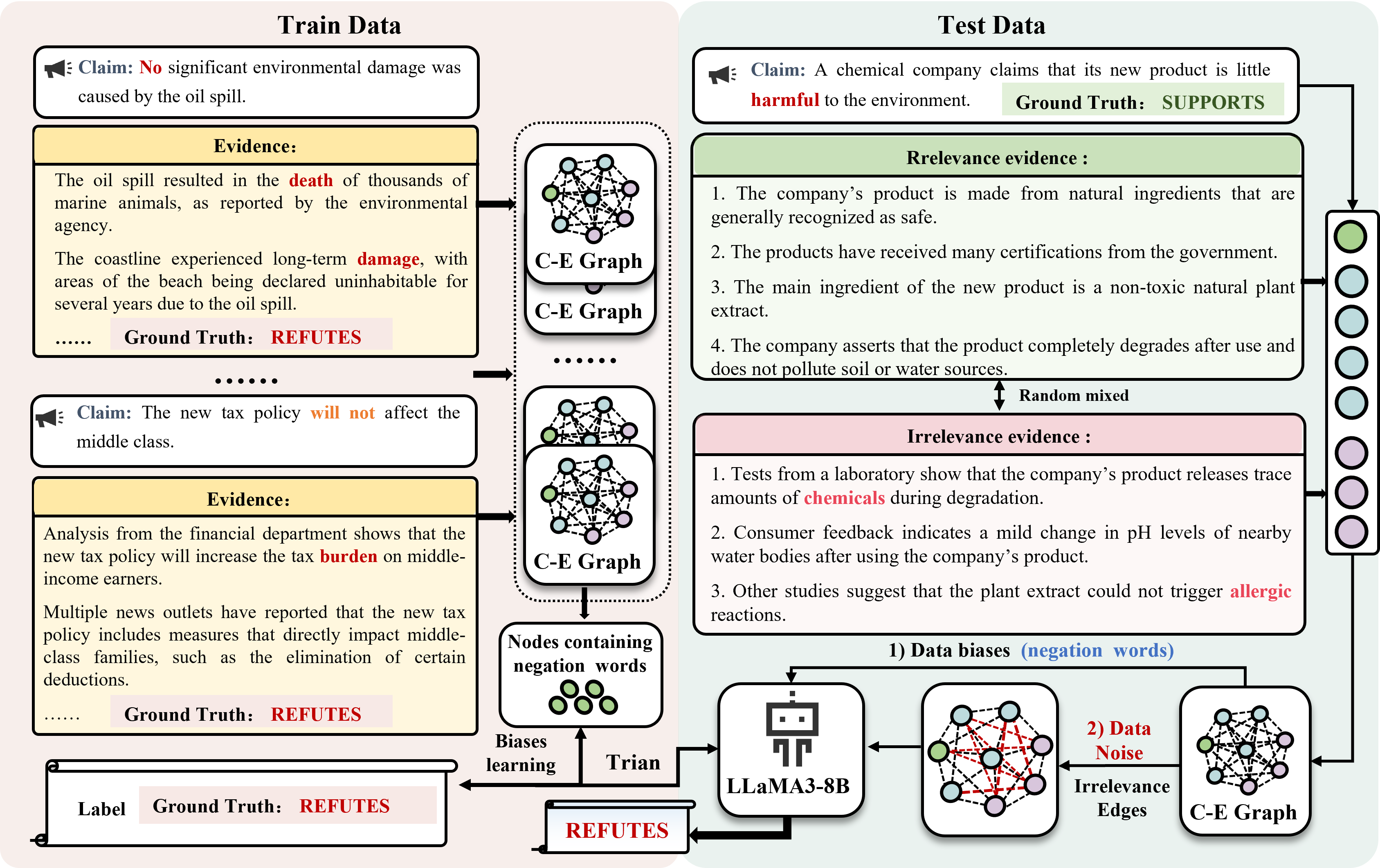}
    \caption{Construction of the C-E Graph and the issues of data biases and noise. Red dashed lines indicate irrelevant edges.
    }
    \label{fig:intro}
\end{figure}

To address traditional methods' limitations, a straightforward approach is to construct a fully connected graph,
denoted as a 
Claim-Evidence Graph (C-E Graph, in short), as shown in Figure.\ref{fig:intro}.
The nodes in C-E Graph represent the claim or each individual piece of evidence, and the edges capture the interactions between the claim and the evidence. 
The construction of the C-E Graph effectively reduces neglect of relationships between evidence.
However, the C-E Graph is often susceptible to interference from confounding such as \textit{data biases} and \textit{data noise}. 

First, in the C-E Graph, the claim and evidential nodes (each node represents the claim and evidence content) are typically collected through manual keyword-based retrieval, which inevitably introduces a significant amount of data bias subjectively.
Subjective bias significantly impacts models with relatively small parameter sizes (e.g., $<$= 10B)  making these models more susceptible to data biases during training, which can affect model performance.
Previous research~\cite{schuster2019towards,zhang2024causal, li2026capturing} reveals that the label REFUTE is highly correlated with negation words or specific statements because of the manual collection process, which may not always hold.
For example, The appearance of the words \textit{``not''}, \textit{``nothing''}, \textit{``nowhere''}, \textit{etc.} are highly correlated with the \textit{``REFUTE''}. Additionally, \textit{``The post was flagged as part''} with \textit{``REFUTE''} label appears in pairs multiple times in the Politihop dataset. 
As a result, the model tends to overly rely on surface-level features and resort to ``shortcut'' reasoning paths \cite{li2020survey}. 
When these keywords are absent, the model’s reasoning ability sharply declines, leading to the so-called ``power outage reasoning'' phenomenon investigated in \cite{xu2021edge}. Current methods (such as data augmentation \cite{wei2019eda, wang2023dual} and weight regularization \cite{schuster2019towards, meng2025heterogeneous}) are often ineffective in eliminating subjective biases in complex reasoning scenarios. 
Therefore, a major research challenge is how to identify and reduce the negative impact caused by data bias introduced by the evidential nodes in the evidence graph.

Secondly, a fully connected C-E Graph that has not been refined often introduces a lot of irrelevant evidence (i.e., redundant evidential nodes) and meaningless evidence related information (i.e., redundant evidence edges), thus causing the problem of data noise.
Data noise often stands out due to its prominent surface features (e.g.,  the terms that appear to look more scientific or seemingly more directly related to the claim) \cite{li2021learning, huang2026video}, thereby diverting the model's attention away from evidence that is truly relevant to the claim and distorting its judgment. While the model may select valid evidence, the significance of this evidence is overshadowed by the noisy data, causing the model to focus predominantly on the one-sided representation of the noisy data. This overemphasis on noisy data, while neglecting valid evidence, leads to the formation of ``information silos'' in the claim verification process, as discussed in \cite{shahrokni2015beyond}. 
For instance, noisy evidence and valid evidence often become entangled in the C-E Graph, disrupting the verification process. As detailed in Figure.\ref{fig:intro}, while \textit{``recognized as safe''} in Evidence 1 is directly relevant to the claim, terms like \textit{``trace chemical releases''} or \textit{``minor pH fluctuations''} in Evidence 4 and 6 are irrelevant. The abundance of irrelevant edges amplifies this noise, obscuring meaningful relationships and diverting the model's attention away from critical evidence.
In summary, many evidence nodes are complexly intertwined, which seriously aggravates the difficulty of verifying the claim.
Therefore, another challenge
is to identify the most relevant evidence from the C-E Graph to improve the verifying ability.

To solve the \textit{\textbf{data biases}} and \textit{\textbf{data noise}} challenges encountered by the fully connected C-E Graph in claim verification, we present a novel method in this paper, named \textbf{Mu}lti-\textbf{P}ath Causa\textbf{l} Optimizati\textbf{on} (\textbf{MuPlon}). More concretely, we leverage the back-door path to dilute the influence of noisy nodes in the C-E Graph while employing local graph augment combined with graph neural networks to strengthen the connections between relevant evidence nodes. Using the front-door path, we extract evidence \textit{subgraphs} from the C-E Graph to construct \textit{reasoning paths} and apply counterfactual reasoning to mitigate biases in these paths. By integrating causal interventions via both \textit{front-door path} and \textit{back-door path}, our approach significantly improves the accuracy and robustness of verification, offering a novel solution to claim verification.
The main contributions of our work are as follows:

\begin{itemize}

\item We propose a novel method, named \textbf{Mu}lti-\textbf{P}ath Causa\textbf{l} Optimizati\textbf{on} (\textbf{MuPlon}), which effectively integrates the \textit{front-door path} and \textit{back-door path} of causal intervention strategies. This method mitigates \textit{\textbf{confounding issues}} related to \textit{\textbf{data noise}} and \textit{\textbf{biases}}, significantly enhancing the accuracy and robustness of claim verification. To the best of our knowledge, we are the first to present a dual causal intervention strategy in the claim verification.

\item In the \textit{back-door path}, we combine a Bayesian network with inverse probability weighting to effectively dilute the influence of noisy nodes in the C-E Graph, while employing a local graph augmentation strategy to alleviate the over-smoothing issue in GNN training, thereby strengthening the connections between evidence nodes. 

\item In the \textit{front-door path}, we design a weight-optimized Markov chain approach to extract subgraphs with strong associations from the C-E Graph to construct high-quality reasoning paths and apply counterfactual reasoning to remove biases in these paths, achieving a more efficient and accurate claim verification.

\item We conducted extensive experiments to compare MuPlon across multiple datasets. Our experiments demonstrate that MuPlon outperforms various previous claim verification methods, achieving state-of-the-art (SOTA) performance and surpassing several large language models in accuracy, further showcasing its advantages in claim verification. These findings further demonstrate MuPlon's scalability and adaptability in data quality control systems for engineering applications.

\end{itemize}

\section{Related Works and Preliminary}

Traditional methods often treat the claim verification task as a natural language inference problem, where models classify evidence as supporting or contradicting a claim based on lexical overlap or syntactic features \cite{rudinger2020thinking}. Earlier claim verification methods relied heavily on such surface-level features that resulted in poor performance.

Subsequently, graph-based models were introduced to better capture the relationships between claims and multi-evidence \cite{nickel2015review}: among them, we cite GEAR \cite{zhou2019gear}, DREAM \cite{zhong2019reasoning} and KGAT \cite{liu2019fine}. However, these approaches often fail to address two key \textit{confounding factors}: \textit{data noise} and \textit{data biases}. Data biases can cause the model to inadvertently learn patterns from imbalanced datasets or overrepresented keywords, leading to skewed predictions, while data noise introduces erroneous or irrelevant information that can interfere with the model's decision-making process.

To remove data biases, recent research has focused on methods like \textit{data augmentation}, \textit{weight regularization}, and \textit{inference intervention} \cite{wei2021model}. CrossAug \cite{lee2021crossaug} generates new training examples by replacing words. CLEVER \cite{xu2023counterfactual} applies counterfactual reasoning to isolate unbiased elements. Causal Walk \cite{zhang2024causal} incorporates mediating variables to address hidden biases. Despite these efforts, these methods handles only shallow bias patterns and struggles with complex tasks.


Furthermore, despite advancements in addressing data biases, a significant open challenge remains in dealing with noisy data, such as contradictory or irrelevant evidence \cite{frenay2013classification}. Although some approaches aim to improve evidence selection or filter irrelevant data \cite{gowda2021smart}, noisy data are a major barrier to achieving reliable and robust claim verification, especially when facing a large number of evidence.

Causal inference, with its robust theoretical foundations and generalization capabilities, has become a popular approach for addressing confounding factors \cite{zelikman2022star,guo2025fgrcat}. By uncovering causal relationships, it provides a solid basis for complex tasks like claim verification. Motivated by this, we first explore structural causal models.

\paragraph{Structural Causal Model}

\begin{figure}[!t]
    \begin{subfigure}{0.16\linewidth}
        \includegraphics[height=1.7cm]{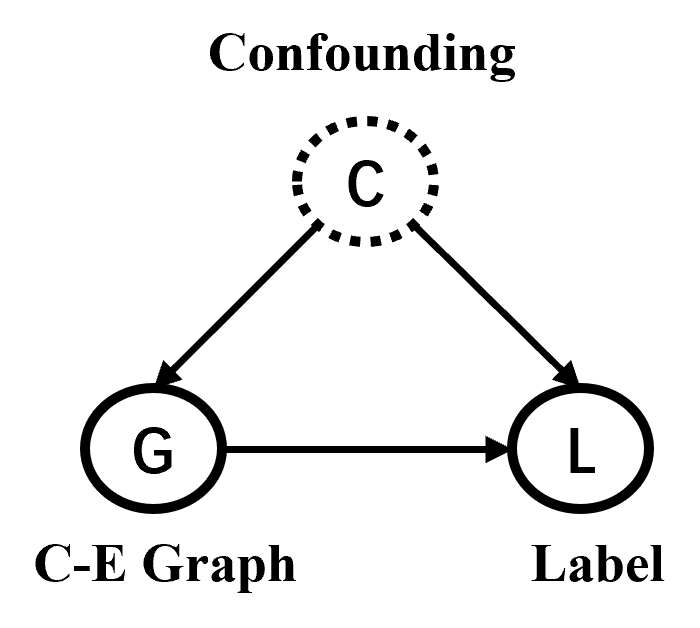} 
        \caption{SCM}
        \label{fig:SCM}
    \end{subfigure}
    \hspace{0.1\linewidth} 
    \begin{subfigure}{0.16\linewidth}
        \includegraphics[height=1.7cm]{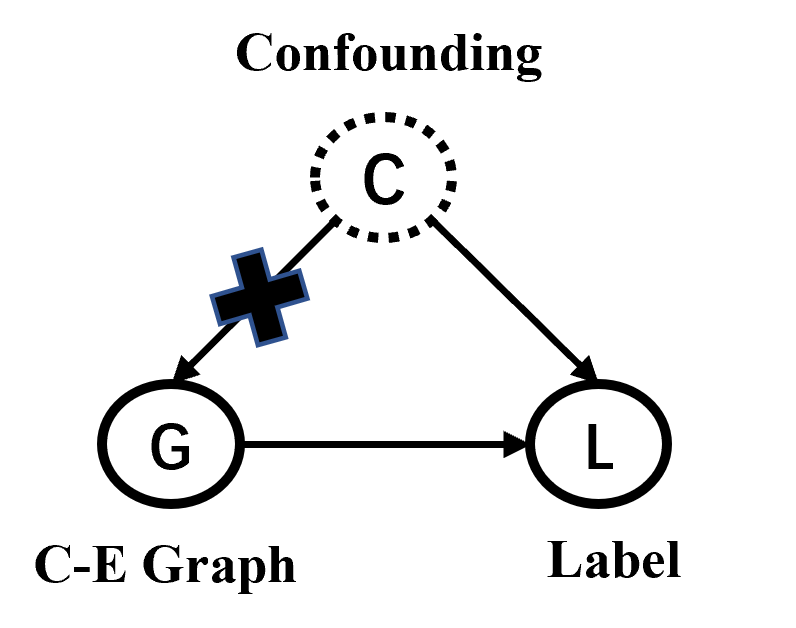} 
        \caption{Back}
        \label{fig:Back}
    \end{subfigure}
    \hspace{0.1\linewidth} 
    \begin{subfigure}{0.16\linewidth}
        \includegraphics[height=1.7cm]{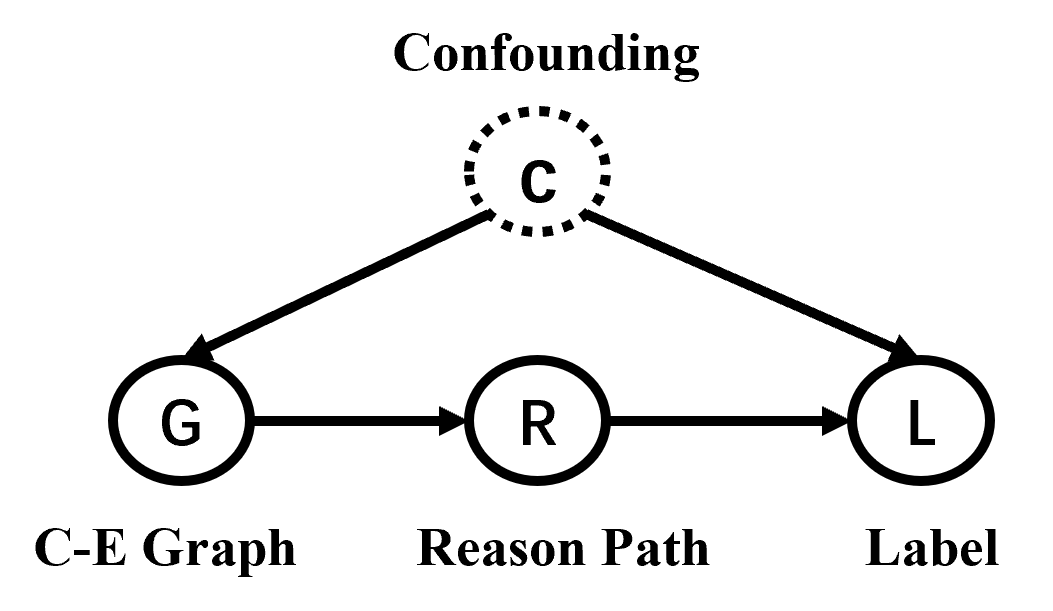} 
        \caption{Front}
        \label{fig:Front}
    \end{subfigure}
    \caption{Structural Causal Model with Causal Intervention.}
    \label{fig:CausalModel}
\end{figure}

We define the Claim-Evidence Graph (C-E Graph, in short) as the variable \(G\), the label of verification as \(L\), and any potential confounding (Data noise and biases) present during inference as \(C\) (see Fig. \ref{fig:SCM}). 

In traditional methods, the total effect \(P(L \mid G)\) measures the influence of input \(G\) on output \(L\). However, when a confounder \(C\) affects \(G\) and \(L\), the direct effect \(P(L \mid G)\) is not accurate and does not account for the spurious correlations introduced by \(C\). Expressed as: \(P(L \mid G) \neq P(L \mid G, C).\)

To address these challenges, Structural Causal Models (SCMs) use \textit{directed acyclic graphs} (DAGs) \cite{keidar2021all} to represent variable interactions and compute causal effects via the \textit{do}-operator. Unlike correlations, the interventional effect \(P(L\mid\text{do}(G))\) isolates the impact of setting \(G\) to a specific value, effectively removing confounding influences of \(C\). This enables SCMs to answer critical questions: \textit{``What would happen to \(L\) if we directly control \(G\)?''}

Key methods for applying \textit{do}-calculus include \textit{back-door path}, \textit{front-door path}, \textit{randomized controlled trials}, and \textit{instrumental variable estimation}. In claim verification, randomized trials and instrumental variables are often infeasible due to limited control over input-output relationships \cite{zhang2024causal}. Thus, back-door path and front-door path adjustment are practical and effective alternatives.

\paragraph{Back-door Path}

As shown in Fig.\ref{fig:Back}, when \(C\) influences \(G\) and \(L\), we can eliminate this influence by back-door adjusting for the confounding. To perform back-door path, we have to identify and control for the confounding variable \(C\) (eg., data noise) to accurately estimate the causal effect of the independent variable \(G\) on the dependent variable \(L\). 

Specifically, the confounding variable \(C\) must satisfy the following conditions: \(C\) affects the relationship between \(G\) and \(L\) and does not have any omitted paths influencing \(L\) from \(C\). For claim verification, we select the controlled confounding (eg., data noise) in the C-E Graph as the confounding variable \(C\). By controlling for \(C\), we can eliminate confounding and correctly estimate \(P(L \mid \text{do}(G))\). We can decompose the causal effect as follows through the Law of Total Probability:
\begin{equation}
P(L|\text{do}(G)) = \sum_{c} P(L|G, c) P(c),
\end{equation}
where \( c \in C \) represents different values of \( C \). We need to compute \( P(L \mid G, c) \) under different conditions of \( c \) and weight them by the marginal probability \( P(c) \) of \( C \).

By controlling for \( C \), we block the back-door paths, ensuring that we can accurately estimate the causal effect of \( G \) on \( L \) as \( P(L \mid do(G)) \), free from the influence of confounding.

\paragraph{Front-door Path}

If data biases of $C$ cause $G$ to have a direct impact on $L$ (see Fig. \ref{fig:Front}) then it is convenient to use a \textit{front-door path}. Front-door path requires the presence of a mediator variable \(R\) that can fully block the causal effect between \(G\) and \(L\).
The reasoning path serves as the mediator variable \(R\). Then, the only pathway through which \(G\) influences \(L\) is the \(R\). Therefore, we can specify the causal pathway as \(G \to R \to L\), \(r \in R\). Thus, we can calculate:
\begin{equation}
P(L|\text{do}(G)) = \sum_{r} P(L|\text{do}(r)) P(r|\text{do}(G)),
\end{equation}
the adjustment decomposes the causal effect between \( G \) and \( L \) into the influence of \( G \) on the mediator variable \( R \) and the influence of \( R \) on \( L \). As \( R \) serves as the mediator, it blocks the direct influence of \( G \) on \( L \), meaning \( G \) can only affect \( L \) indirectly through \( R \), mitigating confounding of \( G \) on \( L \).



\begin{figure*}[!t]
    \centering
    \includegraphics[width=\textwidth]{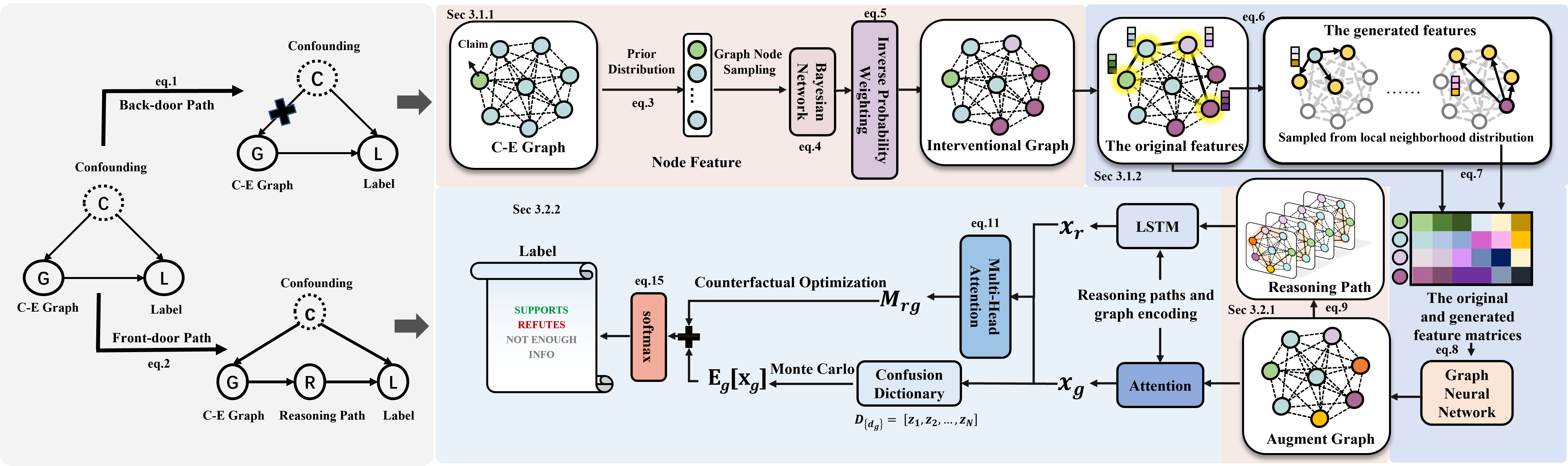}
    \caption{
    An overview of MuPlon Framework (The gray part is the causal intervention theory framework). 
    }
    \label{fig:MuPlon}
\end{figure*}

\section{Proposed Framework: MuPlon}
Since there is no explicitly relationship, we construct the claim and evidence as a fully connected graph (see Appendix for more information), referred to as the Claim-Evidence Graph (C-E Graph), and define claim verification as a graph classification task.
The architecture of MuPlon is graphically reported in Fig. \ref{fig:MuPlon}. MuPlon integrates the \textbf{\textit{Back-door path Adjustment}} with the \textbf{\textit{Front-door path Adjustment}}, aims to enhance the model’s capabilities in claim verification through mitigating Data noise and Data biases ($P(L|\text{do}(G))$).


\subsection{Back-door path Adjustment}

In this subsection, we will detail how to handle \textbf{noisy data}, focusing on two main aspects: First, reducing the influence of evidence nodes that are unrelated to the claim (Noisy node); Second, strengthening the connections between relevant evidence nodes while minimizing the interference from conflicting evidence nodes, ensuring that the model focuses on consistent and compelling evidence combinations.

\subsubsection{Node-Graph Sampling Bayesian Network}

Following previous work \cite{zhang2024causal}, we employ BERT to obtain semantic representations \(X_c\) and \(X_{e}^i\) of the claim \(c\) and evidence \(e_i\) nodes.
Meanwhile, we calculate the attention values of \((e_i, c)\) to indicate the relevance of the evidence to the claim and also compute the semantic similarity of \((e_i, e_j)\) to evaluate the relevance among pieces of evidence. 


Next, we normalize each evidence node and compute its prior probability \(X_{e}^{i^\prime}\), which is integrated with the original node features to enrich the overall representation, as shown in the equation (\( N_e\) is the number of evidence nodes):
\begin{equation}
\begin{split}
X_{e}^{i^\prime} = \frac{X_{e}^i \cdot X_c}{\|X_{e}^i\| \|X_c\|} + \frac{1}{N_e-1} \sum_{\substack{j=1 \\ j \neq i}}^{N_e} \text{Attn}(X_{e}^i, X_{e}^j) \\
\hat{X}_{e}^{i^\prime} = \textstyle \frac{X_{e}^{i^\prime}}{\sum_{j=1}^{N_e} X_{e}^{j^\prime}} \quad i = 1, \ldots, N_e , \quad X_{e}^{i} = \begin{bmatrix} X_{e}^{i} , \hat{X}_{e}^{i^\prime} \end{bmatrix}.
\end{split}
\end{equation}
However, if the number of nodes is too large, distinguishing the differences in prior probabilities after normalization becomes difficult.
This makes it very difficult to distinguish between nodes affected by noise and others. So after obtaining the prior probabilities, we adopt a Bayesian network to update the weight probabilities of the nodes $\hat{X}_{e}^{i^\prime}$.

The Bayesian network is a probabilistic graphical model, representing nodes as random variables and edges as dependencies. A Bayesian network captures uncertainty and associations among nodes, which helps identify noisy data.

In traditional Bayesian graph computations, global inference can pose computational challenges when the graph size is large. To reduce computational complexity, we present a node-based Bayesian graph sampling strategy. 
Specifically, we initiate sampling from nodes with the highest initial weights, updating them and their neighbors. Sampling ceases if the weight variance of all updated nodes remains unchanged for $K_{iter}$ consecutive iterations. This approach hastens convergence and reduces redundant computations by omitting invalid updates. $K_{iter}$ is determined based on the number of graph nodes $N_g$, Mean $\alpha$ and Variance $\beta$ of \(\hat{X}_{e}^{i^\prime}\): \(K_{iter} = \alpha \times \left( \log_2(N_g + 1) + {N_g}(\beta + N_g)^{-1} \right)\).

After determining \( K_{iter} \), the Bayesian network uses conditional probabilities to model dependencies among nodes and calculates the influence of noisy factors through posterior probabilities. Noisy nodes are typically placed at higher levels in a Bayesian network, where they represent a shared influence on multiple downstream nodes. 

Due to this hierarchical structure, changes in higher-level noisy nodes propagate through conditional dependencies, thus amplifying the overall impact of the noise on the network. So, noisy nodes have a greater influence (weight) than other nodes, as defined by the following equation:
\begin{equation}
P(Z \mid \hat{X}_{e}^{i^\prime},\!\text{Pa}(\hat{X}_{e}^{i^\prime})) = 
\tfrac{P(\hat{X}_{e}^{i^\prime} \mid Z,\text{Pa}(\hat{X}_{e}^{i^\prime})) \, 
      P(Z \mid \text{Pa}(\hat{X}_{e}^{i^\prime}))}
     {P(\hat{X}_{e}^{i^\prime} \mid \text{Pa}(\hat{X}_{e}^{i^\prime}))},
\end{equation}
where $\hat{X}_{e}^{i^\prime}$ is a node representing prior probability, $\text{Pa}(\hat{X}_{e}^{i^\prime})$ denotes its parent node set, and \(P(Z \mid \hat{X}_{e}^{i^\prime})\) is the posterior update for noisy factor \(Z\) after observing $\hat{X}_{e}^{i^\prime}$.

To mitigate the influence of noisy factor \(Z\) on inference, we adopt inverse probability weighting to reallocate the probability weights of the nodes \( \tilde{P}(\hat{X}_{e}^{i^\prime}) \), in the equation below:
\begin{equation}
\tilde{P}(\hat{X}_{e}^{i^\prime}) = P(\hat{X}_{e}^{i^\prime}) \cdot P(Z \mid \hat{X}_{e}^{i^\prime})^{-1}.
\end{equation}

In this way, we can ``dilute'' the weight of the noisy node, minimizing their interference with the verification process.

\subsubsection{Local Graph Feature Augmentation}

After diluting noisy data, we use a Graph Neural Network (GNN) to propagate and update node features, strengthening evidence associations among neighboring nodes. This helps the model identify relevant evidence with claim and avoid selecting contradictory evidence.

However, in a fully connected graph, information propagates from a node to its neighbors and, as the number of convolutional layers increases, the vector representing each node tends to converge to the same vector.
This phenomenon, known as ``over-smoothing'', complicates the differentiation between nodes. As a result, the unique characteristics of each node are diminished, leading to a significant loss of differentiation capability among node representations.

To address over-smoothing, we present a \textit{local graph feature augmentation} strategy to preserve the diversity and distinctiveness of node features. Specifically, we use a weighted averaging method based on neighbor weights to update node representation, and, thus, our approach can encode discrepancies between node features as well as the topological structure of the neighborhood of a node. In this way, we avoid the over-smoothing of node features. More formally,

we define a feature generation model that employs a latent variable \(h\) to explain and obtain the relationship between two types of nodes \(X_{e}^{i}\) (the central node) and \(\ X_{e}^j\ _{j \neq i, j \in \{1, 2, \dots, N_t\}}
\) (we briefly define as all neighboring nodes of \(X_{e}^{i}\)). To infer \( h \) from observed data, we must compute the posterior probabilities \( p_{\theta}(h \mid X_e^i, X_e^j) \) for all possible values of \( h \) parametrized by a vector parameter $\theta$. In a fully connected graph with a large number of nodes, due to the large number of neighboring nodes, the exact computation of posterior probabilities is hard, as it involves high-dimensional distributions with no closed-form solution and intractable integrals.



We employ \textit{variational inference} \cite{murphy2012machine} to make the computation of posterior probabilities tractable. Specifically, we propose a tractable variational distribution \(q_{\phi}(h \mid X_{e}^{i}, X_{e}^{j})\) to approximate the true posterior distribution. To achieve this, we maximize the \textit{evidence lower bound}, which measures how well the variational distribution approximates the true posterior. 
The objective function for maximizing the evidence lower bound is as follows:
\begin{equation}
\begin{split}
\mathcal{L}(X_{e}^{i}, X_{e}^{j}; \theta, \phi) = &\mathcal{W}_1(q_\phi(h|X_{e}^{i}, X_{e}^{j}) \| p_\theta(h|X_{e}^{j})) \\
& + \mathbb{E}_{q_\phi(h|X_{e}^{i}, X_{e}^{j})}[\log p_\theta(X_{e}^{i}|X_{e}^{j}, h)],
\end{split} 
\end{equation}
where \( \mathcal{W}_1 \) is the Wasserstein-1, quantifying the minimal cost to align two distributions in high-dimensional space. 
The prior distribution \( p_\theta(h \mid X_{e}^{j}) \) represents the latent variable \( h \) for \( X_{e}^{j} \), providing an initial hypothesis and structural information. This prior, combined with the optimizing parameters from the \( \mathcal{W}_1 \) , allows the variational distribution \( q_\phi(h \mid X_{e}^{i}, X_{e}^{j}) \) to effectively approximate the posterior distribution \( p(h \mid X_{e}^{i}, X_{e}^{j}) \), the conditional probability of the generative model \( p_\theta(X_{e}^{i} \mid X_{e}^{j}, h) \) describes the probability of generating the features of node \( X_{e}^{i} \) given its neighbors \( X_{e}^{j} \) and the latent variable \( h \).
The optimization of $\mathcal{L}(X_{e}^{i}, X_{e}^{j}; \theta, \phi)$ comprises two steps:
\textbf{\textit{a)}} \textbf{Optimizing $\phi$}: This allows the variational distribution $q_{\phi}(h \mid X_{e}^{i}, X_{e}^{j})$ to better approximate the true posterior, improving the inference of the latent variable $h$.
\textbf{\textit{b)}} \textbf{Optimizing $\theta$}: This enhances the conditional distribution $p_{\theta}(X_{e}^{i} \mid X_{e}^{j}, h)$, improving the model's ability to generate or reconstruct the node features.
Optimizing $\phi$ and $\theta$ jointly enhances model performance on complex graph data.

After solving the optimization problem above, we can sample \(h\) from the variational distribution and use it to infer and generate the features \(X^i_{\text{generated}}\), represented as:
\begin{equation}
h \sim q_{\phi^*}(h | X_{e}^{i}, X_{e}^{j}), \quad X^i_{\text{generated}} \sim p_{\theta^*}(X | X_{e}^{j}, h).
\end{equation}

Next, we combine the generated features \(X^i_{\text{generated}}\) with the original node information features, then processed through GNN to boost node feature expressiveness:
\begin{equation}
\begin{split}
X^{l_{k-1}} &= \sum_{i=1}^{N_e} \left( w_1 X^{l_{k-1}}[i,:] + w_2 X^i_{\text{generated}} \right) \\
X^{l_k} &= \text{GNN}(X^{l_{k-1}}, A),
\end{split}
\end{equation}
where \( X^{l_k} \) and \( X^{l_{k-1}} \) denote the node feature matrices for the current and previous layers, respectively. For each node \( i \), its feature is updated by combining its previous layer feature \( X^{l_{k-1}}[i,:] \) with the generated feature \( X^i_{\text{generated}} \).



Our local graph feature augmentation strategy enables us to effectively capture the neighborhood structure of nodes, leading to two key benefits: \textit{a)} avoiding over-smoothing, and \textit{b)} strengthening connections between different types of evidence nodes within the C-E Graph.

\subsection{Front-door path Adjustment}

\textbf{Data biases} often cause the model to learn superficial features. To address it, we propose using the front-door path. Considering the front-door path criterion, mediators need to be introduced. Meanwhile, redundancy in the full graph introduces unnecessary information, hindering inference. Therefore, we construct reasoning paths (key info) from the graph, using them as mediators to address biases. Specifically:


\subsubsection{Weighted Optimization Markov Chain Path Selection}

Global path exploration during inference requires searching all possible paths, increasing computational complexity. In large-scale graph scenarios, it can lead to path explosion, complicating the identification of the optimal path.



To address these challenges, we propose a \textit{Markov chain search strategy} based on \textit{weight optimization}. This method relies solely on the current node state for transitions and, in this way, it avoids the high costs of global search. Specifically:
for any pair of adjacent nodes $(x_i, x_j)$ in the graph, we apply a multi-layer perceptron (MLP) to compute initial transition weights: \(a_{ij} = \text{MLP}(x_i, x_j)\).


We adjust the transition weight \( \tilde{a}_{ij} \) for each node to mitigate the influence of noisy factors by incorporating inverse probability weights in transition weights: \(\tilde{a}_{ij} = a_{ij} \cdot \tilde{P}(\hat{X}_{e}^{i^\prime})\).   



This adjustment ensures that the state transition weights depend solely on the current node's information. Subsequently, we calculate the transition probability from \( v_i \) to \( v_j \):
\begin{equation}
P(j \mid i) = \text{softmax}(\tilde{a}_{ij}) = \frac{\exp\left(\tilde{a}_{ij}\right)}{\sum_{k \in N(i)} \exp\left(\tilde{a}_{ik}\right)}.
\end{equation}





Since there may be multiple potential paths, we first calculate the product score of node transition probabilities for each path as a path score. Then, using beam search, we retain the top five candidate paths which are high scores, increasing the efficiency of subsequent reasoning.

\subsubsection{Counterfactual Removal of Graph Bias}
 


Since training sets introduce data biases that limit the model's performance, we aim to remove bias from reasoning paths. 

However, directly using the dataset to calculate bias is challenging due to the high dimensionality and complex dependencies within the data. 
To achieve our goal, we introduce \textit{counterfactual optimization}. Specifically, we represent each data sample in the training set as a collection of C-E Graph and construct a \textit{confusion dictionary} to integrate all bias' features. This allows us to analytically model biases introduced by the dataset and compute the expected bias of the C-E Graph. By combining path and graph features and subtracting the expected bias, we eliminate the interference caused by data biases, enhancing the model’s inference ability.

Specifically:
to capture both sequential and long-range dependencies within the reasoning path, we employ LSTM to encode the path representation \( x_r \). Simultaneously, we utilize an attention mechanism to learn the augmented graph \( x_g \).

After obtaining \(x_r\) and \(x_g\), we apply a multi-head attention mechanism to fuse these representations, mitigating potential global information loss in the path representation extracted via beam search. This mechanism maps \(x_r\) and \(x_g\) into subspaces to learn intricate relationships efficiently. For the \(i\)-th attention head, we compute the query \(Q_r^i\), key \(K_g^i\), and value \(V_g^i\) using linear transformations:
\begin{equation}
Q_r^i = W_q^i x_r, \quad K_g^i = W_k^i x_g, \quad V_g^i = W_v^i x_g,  
\end{equation}
where \(W_{q/k/v}^i\) are the learnable parameters.
Dot-product attention computes scores (path and graph), producing \(\text{head}^i\) for each head. These outputs are concatenated and linearly transformed to yield the fused representation:
\begin{equation}
M_{(rg)} = \text{Z}(x_r, x_g) = W_o \cdot \text{Concat}(\text{head}^1, \ldots, \text{head}^H)
\end{equation}


Since the paths are part of the graph, the bias in the paths often originates from the overall bias of the graph. Therefore, we adopt a ``baseline correction'' based on counterfactual reasoning to neutralize the bias value of the graph between the path and graph attention values. In this way, the path can accurately represent the unbiased information in the graph.




The calculation of the bias expected value of $x_g$ by means of an equation is often unfeasible due to the graph's nonlinear characteristics and inherent randomness. To address the uncertainties in complex graph structures, we construct a \( D_{d_g} \) which encompasses all relevant bias features observed in the training set. Then, we adopt a Monte Carlo algorithm \cite{rubinstein2016simulation} to obtain bias expected vector \( \mathbb{E}[x_g] \).

Specifically, for each graph sample feature representation \( x_{g}' \) obtained from the training set, we adopt the K-Means clustering algorithm to partition the samples from each category into multiple cluster centers. The cluster centers for each category can be expressed as:
\begin{equation}
    z_i = [z_{i1}, z_{i2}, \ldots, z_{ik}] \in \mathbb{R}^{k \times d},
\end{equation}
here, \( z_{ij} \) represents the \( j \)-th cluster center of category \( l_i \), obtained from clustering each \( x_{g}' \). The variable \( k \) denotes the number of cluster centers per category (defined as twice the number of categories (\(N\)), while \( d \) is the feature dimension. The cluster center \( z_i \) represents the feature distribution of samples within that category.

Combining all the cluster centers from different categories forms the confusion dictionary \( D_{d_g} \):
\begin{equation}
D_{d_g} = [z_1, z_2, \ldots, z_N] \in \mathbb{R}^{N \times k \times d},    
\end{equation}
here, \( N \) is the number of categories, \( k \) is the number of cluster centers for each category, and \( d \) is the feature dimension.


After constructing the confusion dictionary \( D_{d_g} \), we use the Monte Carlo algorithm to estimate vector \( \mathbb{E}[x_g] \). We randomly select \( M \) cluster centers \( \{ z_1, z_2, \ldots, z_M \} \) from \( D_{d_g} \), where \( M = \left\lceil {N}/{2} \right\rceil \). For each cluster center \( z_i \), we calculate its corresponding graph representation \( x_{g_i} \) and use these samples to estimate the expected value of \( x_g \). The steps are:
\begin{equation}
x_{g_i} = f(z_i), \quad \mathbb{E}[x_g] \approx \frac{1}{M} \sum_{i=1}^{M} x_{g_i},
\end{equation}
where \( f() \) is the feature fusion function that maps the cluster centers to the graph representation.

Thus, after obtaining the biases expected vector, we can eliminate the data biases from the reasoning path, and input the results into the classifier to obtain the final prediction \(\hat{L}\):
\begin{equation}
\hat{L} = \text{softmax}(M_{(rg)} - \alpha W_g \cdot \mathbb{E}[x_g]),
\end{equation}
where \(\alpha\) is a regularization parameter, \(W_g\) is the weight matrix used to transform and combine input features.

\section{Experiments}
\subsection{Experimental Setups}

\subsubsection{Datasets}

\textbf{\textit{a)}} FEVER \cite{thorne2018fever} is a large claim verification dataset from Wikipedia, with claims labeled as ``SUPPORTS'', ``REFUTES'', or ``NOT ENOUGH INFO''. Adversarial FEVER introduces deceptive samples to test model robustness. FEVER-MH requires integrating multiple evidence snippets for verification. Adversarial FEVER-MH adds complexity to test model stability.
\textbf{\textit{b)}} Politihop \cite{ostrowski2020multi} is designed for multi-hop verification of political claims. Adversarial Politihop contains misleading statements to test the model’s ability to handle inconsistent information. Politihop-Adv2 is binary, Politihop-Adv3 is three-class. Hard Politihop \cite{zhang2024causal} uses GPT-4 to generate complex misleading evidence, and Symmetric Politihop combines GPT-4 generated and original samples to assess debiasing. ShareEvi increases complexity by reusing evidence in new claims.
\textbf{\textit{c)}} Cladder \cite{jin2024cladder} is a step-by-step inference, and evaluates causal inference abilities, covering associative, interventional, and counterfactual. It includes: \textit{Easy}, \textit{Hard}, \textit{Commonsense}, \textit{Anticommonsense}, and \textit{Nonsense}.

\subsubsection{Baselines}

We compare MuPlon with several methods, including  BERT-Concat \cite{zhang2021ma}, KGAT \cite{liu2019fine}, GEAR \cite{zhou2019gear}, RAV \cite{zheng2024evidence}, DREAM \cite{zhong2019reasoning}, CrossAug \cite{lee2021crossaug}, Transformer-XH \cite{zhao2020transformer}, as well as causal methods such as Causal Walk \cite{zhang2024causal}, CICR \cite{tian2022multimodal} and CLEVER \cite{xu2023counterfactual}. Meanwhile, CICR-Graph, and CLEVER-Graph replace the encoders of CICR and CLEVER with GNN \cite{zhang2024causal}. Furthermore, we further compare with Random \cite{jin2024cladder}, language models (Alpaca \cite{maeng2017alpaca}, GPT-3 \cite{ouyang2022training}) to evaluate the performance in the Cladder dataset.

\begin{table*}[ht]
    \centering
    \caption{Accuracy comparison of MuPlon and various methods on the Fever and Politihop datasets.}
    \setlength{\tabcolsep}{7pt}
    \renewcommand{\arraystretch}{1.2} 
    \scalebox{1}{
    \begin{tabular}{c|c|c|c|c|c|c|c}
        \hline
        \textbf{Method} & \textbf{FEVER} & \textbf{FEVER-MH} & \textbf{Advers.-FEVER} & \textbf{Advers.-FEVER-MH} & \textbf{Politihop} & \textbf{Politihop-Adv3} & \textbf{Politihop-Hard} \\
        \hline
        \textbf{MuPlon} & \textbf{91.9} & \textbf{93.3} & \textbf{64.6} & \textbf{68.1} & \textbf{80.5} & \textbf{79.5} & \textbf{79.0} \\
        \hline
        BERT-Concat & 82.1\textsubscript{\textcolor{darkred}{$\downarrow$ 9.8}} & 86.3\textsubscript{\textcolor{darkred}{$\downarrow$ 7.0}} & 59.1\textsubscript{\textcolor{darkred}{$\downarrow$ 5.5}} & 62.1\textsubscript{\textcolor{darkred}{$\downarrow$ 6.0}} & 76.0\textsubscript{\textcolor{darkred}{$\downarrow$ 4.5}} & 74.5\textsubscript{\textcolor{darkred}{$\downarrow$ 5.0}} & 71.5\textsubscript{\textcolor{darkred}{$\downarrow$ 7.5}} \\

        RAV & 88.3\textsubscript{\textcolor{darkred}{$\downarrow$ 3.6}} & 89.6\textsubscript{\textcolor{darkred}{$\downarrow$ 3.7}} & 59.2\textsubscript{\textcolor{darkred}{$\downarrow$ 5.4}} & 64.1\textsubscript{\textcolor{darkred}{$\downarrow$ 4.0}} & - & - & - \\

        GEAR & 86.6\textsubscript{\textcolor{darkred}{$\downarrow$ 5.3}} & 87.0\textsubscript{\textcolor{darkred}{$\downarrow$ 6.3}} & 57.8\textsubscript{\textcolor{darkred}{$\downarrow$ 6.8}} & 63.4\textsubscript{\textcolor{darkred}{$\downarrow$ 4.7}} & 75.5\textsubscript{\textcolor{darkred}{$\downarrow$ 5.0}} & 75.0\textsubscript{\textcolor{darkred}{$\downarrow$ 4.5}} & 73.5\textsubscript{\textcolor{darkred}{$\downarrow$ 5.5}} \\

        DREAM & 76.9\textsubscript{\textcolor{darkred}{$\downarrow$ 15.0}} & - & - & - & - & - & - \\

        KGAT & 86.7\textsubscript{\textcolor{darkred}{$\downarrow$ 5.2}} & 89.9\textsubscript{\textcolor{darkred}{$\downarrow$ 3.4}} & 59.3\textsubscript{\textcolor{darkred}{$\downarrow$ 5.3}} & 64.9\textsubscript{\textcolor{darkred}{$\downarrow$ 3.2}} & 77.0\textsubscript{\textcolor{darkred}{$\downarrow$ 3.5}} & 74.5\textsubscript{\textcolor{darkred}{$\downarrow$ 5.0}} & 74.0\textsubscript{\textcolor{darkred}{$\downarrow$ 5.0}} \\
        
        CrossAug & 86.1\textsubscript{\textcolor{darkred}{$\downarrow$ 5.9}} & - & 52.8\textsubscript{\textcolor{darkred}{$\downarrow$ 11.8}}  & - & - & - & - \\

        Transformer-XH & 83.1\textsubscript{\textcolor{darkred}{$\downarrow$ 8.8}} & 86.6\textsubscript{\textcolor{darkred}{$\downarrow$ 6.7}} & 59.4\textsubscript{\textcolor{darkred}{$\downarrow$ 5.2}} & 64.4\textsubscript{\textcolor{darkred}{$\downarrow$ 3.7}} & 75.5\textsubscript{\textcolor{darkred}{$\downarrow$ 5.0}} & 77.0\textsubscript{\textcolor{darkred}{$\downarrow$ 2.5}} & 72.3\textsubscript{\textcolor{darkred}{$\downarrow$ 6.7}} \\ 

        Causal Walk & 90.2\textsubscript{\textcolor{darkred}{$\downarrow$ 1.7}} & 92.9\textsubscript{\textcolor{darkred}{$\downarrow$ 0.4}} & 62.1\textsubscript{\textcolor{darkred}{$\downarrow$ 2.5}} & 67.1\textsubscript{\textcolor{darkred}{$\downarrow$ 1.0}} & 80.0\textsubscript{\textcolor{darkred}{$\downarrow$ 0.5}} & 79.0\textsubscript{\textcolor{darkred}{$\downarrow$ 0.5}} & 79.0\textsubscript{\textcolor{darkred}{$\downarrow$ 0.0}} \\

        CICR & 79.3\textsubscript{\textcolor{darkred}{$\downarrow$ 12.6}} & 83.4\textsubscript{\textcolor{darkred}{$\downarrow$ 9.9}} & 61.9\textsubscript{\textcolor{darkred}{$\downarrow$ 2.7}} & 64.1\textsubscript{\textcolor{darkred}{$\downarrow$ 5.0}} & 76.0\textsubscript{\textcolor{darkred}{$\downarrow$ 4.5}} & 74.5\textsubscript{\textcolor{darkred}{$\downarrow$ 5.0}} & 75.0\textsubscript{\textcolor{darkred}{$\downarrow$ 4.0}} \\

        CLEVER & 78.7\textsubscript{\textcolor{darkred}{$\downarrow$ 13.2}} & 82.5\textsubscript{\textcolor{darkred}{$\downarrow$ 10.8}} & 59.9\textsubscript{\textcolor{darkred}{$\downarrow$ 4.7}} & 64.4\textsubscript{\textcolor{darkred}{$\downarrow$ 3.7}} & 76.0\textsubscript{\textcolor{darkred}{$\downarrow$ 4.5}} & 76.0\textsubscript{\textcolor{darkred}{$\downarrow$ 3.5}} & 73.0\textsubscript{\textcolor{darkred}{$\downarrow$ 6.0}} \\

        CICR-Graph & 87.4\textsubscript{\textcolor{darkred}{$\downarrow$ 4.5}} & 91.5\textsubscript{\textcolor{darkred}{$\downarrow$ 1.8}} & 59.2\textsubscript{\textcolor{darkred}{$\downarrow$ 5.4}} & 65.8\textsubscript{\textcolor{darkred}{$\downarrow$ 2.3}} & 78.0\textsubscript{\textcolor{darkred}{$\downarrow$ 2.5}} & 77.5\textsubscript{\textcolor{darkred}{$\downarrow$ 2.0}} & 76.5\textsubscript{\textcolor{darkred}{$\downarrow$ 2.5}} \\

        CLEVER-Graph & 86.2\textsubscript{\textcolor{darkred}{$\downarrow$ 5.7}} & 89.9\textsubscript{\textcolor{darkred}{$\downarrow$ 3.4}} & 59.5\textsubscript{\textcolor{darkred}{$\downarrow$ 5.1}} & 64.6\textsubscript{\textcolor{darkred}{$\downarrow$ 3.5}} & 78.0\textsubscript{\textcolor{darkred}{$\downarrow$ 2.5}} & 76.5\textsubscript{\textcolor{darkred}{$\downarrow$ 3.0}} & 75.5\textsubscript{\textcolor{darkred}{$\downarrow$ 3.5}} \\
        \hline
    \end{tabular}}
    \label{tab:1}
\end{table*}

\begin{table}[ht]
    \centering
    \caption{Accuracy on Symmetric Politihop dataset.}
    \setlength{\tabcolsep}{3pt} 
    \renewcommand{\arraystretch}{1.2} 
    \resizebox{\linewidth}{!}{
    \begin{tabular}{l|c|c|c}
        \hline
        \textbf{Method} & \textbf{Politihop-Adv2} & \textbf{Symmetric-Politihop-ShareEvi}  & \textbf{Symmetric-Politihop} \\
        \hline
        \textbf{MuPlon} & \textbf{89.6} & \textbf{65.5} & \textbf{61.4} \\
        \hline
        GEAR & 
        89.5\textsubscript{\textcolor{darkred}{$\downarrow$ 0.1}} & 
        50.6\textsubscript{\textcolor{darkred}{$\downarrow$ 14.9}} & 
        51.2\textsubscript{\textcolor{darkred}{$\downarrow$ 10.2}} \\ 

        Causal Walk & 
        88.3\textsubscript{\textcolor{darkred}{$\downarrow$ 1.3}} & 
        57.0\textsubscript{\textcolor{darkred}{$\downarrow$ 8.5}} & 
        54.1\textsubscript{\textcolor{darkred}{$\downarrow$ 7.3}} \\ 

        CICR-Graph & 
        87.7\textsubscript{\textcolor{darkred}{$\downarrow$ 1.9}} & 
        50.6\textsubscript{\textcolor{darkred}{$\downarrow$ 14.9}} & 
        51.8\textsubscript{\textcolor{darkred}{$\downarrow$ 9.6}} \\ 

        CLEVER-Graph & 
        86.6\textsubscript{\textcolor{darkred}{$\downarrow$ 3.0}} & 
        53.2\textsubscript{\textcolor{darkred}{$\downarrow$ 12.3}} & 
        52.1\textsubscript{\textcolor{darkred}{$\downarrow$ 9.3}} \\ 
        \hline
    \end{tabular}}
    \label{tab:2}
\end{table}

\begin{table}[h]
    \centering
    \caption{Accuracy on Cladder Datasets.}
    \setlength{\tabcolsep}{3pt} 
    \resizebox{\linewidth}{!}{
    \begin{tabular}{l|c|c|c|c|c}
      \hline
      Datasets & \textbf{MuPlon} & Causal Walk & Random & Alpaca & GPT-3 \\    
      \hline
      Cladder-anti & \textbf{49.7} & 
      48.6\textsubscript{\textcolor{darkred}{$\downarrow$ 1.1}} & 
      49.7\textsubscript{\textcolor{darkred}{$\downarrow$ 0.0}} & 
      45.9\textsubscript{\textcolor{darkred}{$\downarrow$ 3.8}} & 
      47.0\textsubscript{\textcolor{darkred}{$\downarrow$ 2.7}} \\

      Cladder-det & \textbf{50.0} & 
      50.0\textsubscript{\textcolor{darkred}{$\downarrow$ 0.0}} & 
      49.1\textsubscript{\textcolor{darkred}{$\downarrow$ 0.9}} & 
      45.5\textsubscript{\textcolor{darkred}{$\downarrow$ 4.5}} & 
      47.0\textsubscript{\textcolor{darkred}{$\downarrow$ 3.0}} \\

      Cladder-easy & \textbf{60.7} & 
      48.6\textsubscript{\textcolor{darkred}{$\downarrow$ 12.1}} & 
      50.3\textsubscript{\textcolor{darkred}{$\downarrow$ 10.4}} & 
      61.3\textsubscript{\textcolor{darkgreen}{$\uparrow$ 0.6}} & 
      61.9\textsubscript{\textcolor{darkgreen}{$\uparrow$ 1.2}} \\

      Cladder-hard & \textbf{49.5} & 
      48.4\textsubscript{\textcolor{darkred}{$\downarrow$ 1.1}} & 
      49.1\textsubscript{\textcolor{darkred}{$\downarrow$ 0.4}} & 
      41.9\textsubscript{\textcolor{darkred}{$\downarrow$ 7.6}} & 
      44.9\textsubscript{\textcolor{darkred}{$\downarrow$ 4.6}} \\

      Cladder-non & \textbf{47.6} & 
      47.5\textsubscript{\textcolor{darkred}{$\downarrow$ 0.1}} & 
      49.0\textsubscript{\textcolor{darkgreen}{$\uparrow$ 1.4}} & 
      45.2\textsubscript{\textcolor{darkred}{$\downarrow$ 2.4}} & 
      48.3\textsubscript{\textcolor{darkgreen}{$\uparrow$ 0.7}} \\ 
      \hline
    \end{tabular}}
    \label{tab:3}
\end{table}

\subsubsection{Implementations}

The experiments are conducted on an NVIDIA GeForce RTX 3090 GPU using BERT as the baseline. 
BERT is chosen for its wide applicability and maturity, enabling efficient comparison with prior work while conserving computational resources.
Dataset and experiment details are available in the \textbf{Anonymous Links} in the Abstract (For more experimental comparisons, please refer to the appendix.).


\subsection{Experimental Results}
\subsubsection{Overall Performance}

We compared several methods across different datasets. As shown in TABLE \ref{tab:1}, the key findings are as follows:

\textbf{\textit{a)}} On the FEVER and FEVER-MH datasets, MuPlon achieves accuracies of 91.9\% and 93.3\%, respectively, outperforming other methods such as GEAR (86.6\% and 87.0\%) and KGAT (86.7\% and 89.9\%). Although Causal Walk performs well with 90.2\% and 92.9\% on these datasets, MuPlon outperforms it, showing superior generalization for both single and multi-hop claim verification. 
\textbf{\textit{b)}} In the Adversarial-FEVER and Adversarial-FEVER-MH datasets, MuPlon achieves 64.6\% and 68.1\% accuracy, respectively, surpassing other methods such as KGAT (59.3\% and 64.9\%), CLEVER-Graph (59.5\% and 64.6\%) and Causal Walk (62.1\% and 67.08\%). Overall, MuPlon stands out as the top performer among all methods in adversarial conditions.
\textbf{\textit{c)}} On the Politihop-Hard dataset, MuPlon achieves 79.0\% accuracy, outperforming BERT-Concat (71.5\%) and Transformer-XH (72.3\%). Additionally, MuPlon performs strongly on the Politihop and Politihop-Adv3 datasets, with scores of 80.5\% and 79.5\%, respectively. Although Causal Walk reaches 79.0\% on Politihop-Hard, MuPlon’s consistent performance across all Politihop highlights its superior adaptability, showing its strength in complex claim verification.

\subsubsection{Results on Symmetric Politihop and Cladder Datasets}

TABLE \ref{tab:2} summarizes the performance on the Symmetric Politihop dataset. MuPlon demonstrates robust performance, achieving 89.6\% on Politihop-Adv2, 65.5\% on Symmetric-Politihop-ShareEvi, and 61.4\% on Symmetric-Politihop. In contrast, methods like GEAR and Causal Walk show significant drops on symmetric tasks. GEAR scores 89.5\% on Politihop-Adv2 but falls to 50.6\% and 51.2\% on the symmetric datasets, while Causal Walk drops from 88.3\% to 57.0\% and 54.1\%. CICR-Graph and CLEVER-Graph exhibit even weaker performance, highlighting MuPlon’s robustness.

\begin{table}[!t]
    \centering
    \caption{Ablation experiments (Accuracy) of MuPlon.}
    \setlength{\tabcolsep}{1pt} 
    \renewcommand{\arraystretch}{1.2} 
    \resizebox{\columnwidth}{!}{
    \begin{tabular}{l|c|c|c}
        \hline
        \textbf{Datasets} & \textbf{MuPlon} & \textbf{w/o Back-door Path} & \textbf{w/o Front-door Path} \\
        \hline
        \textbf{FEVER} & \textbf{91.9} & 88.7\textsubscript{\textcolor{darkred}{$\downarrow$ 3.2}} & 89.8\textsubscript{\textcolor{darkred}{$\downarrow$ 2.1}} \\
        \textbf{FEVER-MH} & \textbf{93.3} & 90.7\textsubscript{\textcolor{darkred}{$\downarrow$ 2.6}} & 91.5\textsubscript{\textcolor{darkred}{$\downarrow$ 1.8}} \\
        \textbf{Adversarial-FEVER} & \textbf{64.6} & 61.1\textsubscript{\textcolor{darkred}{$\downarrow$ 3.5}} & 60.3\textsubscript{\textcolor{darkred}{$\downarrow$ 4.2}} \\
        \textbf{Adversarial-FEVER-MH} & \textbf{68.1} & 65.3\textsubscript{\textcolor{darkred}{$\downarrow$ 2.8}} & 64.4\textsubscript{\textcolor{darkred}{$\downarrow$ 3.7}} \\
        \textbf{Politihop} & \textbf{80.5} & 77.5\textsubscript{\textcolor{darkred}{$\downarrow$ 3.0}} & 78.0\textsubscript{\textcolor{darkred}{$\downarrow$ 2.5}} \\
        \textbf{Politihop-Adv3} & \textbf{79.5} & 77.0\textsubscript{\textcolor{darkred}{$\downarrow$ 2.5}} & 77.0\textsubscript{\textcolor{darkred}{$\downarrow$ 2.5}} \\
       \textbf{Politihop-hard} & \textbf{79.0} & 78.0\textsubscript{\textcolor{darkred}{$\downarrow$ 1.0}} & 77.0\textsubscript{\textcolor{darkred}{$\downarrow$ 2.0}} \\
        \hline
    \end{tabular}}
    \label{tab:4}
\end{table}

Table \ref{tab:3} highlights the accuracy of the Cladder datasets. MuPlon demonstrates competitive performance and frequently outperforms Causal Walk. While Alpaca and GPT-3 excel on Cladder-easy, they struggle with more challenging datasets like Cladder-hard. This underscores MuPlon’s ability to match the performance of LLMs on simpler tasks while outperforming them in more complex scenarios.

\subsubsection{Ablation Study}

Table \ref{tab:4} presents the ablation experiments of MuPlon, demonstrating the effectiveness of our proposed framework.

On the one hand, removing the back-door path leads to a drop in accuracy, as the model can no longer effectively differentiate the importance of the claim and evidence nodes. This results in suboptimal path selection for the front-door path, ultimately affecting overall accuracy.

On the other hand, excluding the front-door path increases the complexity of global graph reasoning, potentially compromising accuracy, and limits the model's ability to address data biases effectively, further diminishing accuracy.

\section{Conclusions and Future Works}

In this work, we explore the limitations of current claim verification methods, especially in handling confounding factors like \textbf{\textit{data biases}} and \textbf{\textit{data noise}}. To solve these problems, we innovative propose MuPlon (\textbf{Mu}lti-\textbf{P}ath Causa\textbf{l} Optimizati\textbf{on}), a novel framework combining the back-door path of graph learning to augment features and the front-door path of counterfactual reasoning to optimize paths. Our novel framework outperforms existing methods in accuracy and achieves state-of-the-art (SOTA), refining the reliability of claim verification.
In future work, we aim to further expand MuPlon’s applicability to broader domains and explore its scalability for real-time and step-by-step verification tasks.

\bibliographystyle{IEEEtran}
\bibliography{custom}

\end{document}